\documentclass[aps,pre,twocolumn,showpacs,preprintnumbers,amsmath,amssymb,floatfix]{revtex4-1}
\usepackage{graphicx}
\usepackage{amsfonts}
\usepackage{CJK}
\usepackage{booktabs}
\usepackage{color}
\usepackage{epstopdf}
\usepackage[dvipdfm,colorlinks,urlcolor=blue,linkcolor=blue,anchorcolor=blue,citecolor=blue]{hyperref}

\begin{document}
\bibliographystyle{apsrev4-1}

\title{How and what to learn: the modes of machine learning}

\author{Sihan Feng$^1$}
\author{Yong Zhang$^{1,2}$}
\author{Fuming Wang$^1$}
\author{Hong Zhao$^{1,2}$}
\email{zhaoh@xmu.edu.cn}
\affiliation{$^1$Department of Physics, Xiamen University, Xiamen 361005, China\\
$^2$Lanzhou Center for Theoretical Physics, Key Laboratory of Theoretical Physics of Gansu Province, Lanzhou University, Lanzhou, 730000, China}

\date{\today}

\begin{abstract}

Despite their great success, neural networks still remain as ``black-boxes'' due to the lack of interpretability. Here we propose a new analyzing method, namely the weight pathway analysis (WPA), to make them transparent. We consider weights in pathways that link neurons longitudinally from input neurons to output neurons, or simply weight pathways, as the basic units for understanding a neural network, and decompose a neural network into a series of subnetworks of such weight pathways. A visualization scheme of the subnetworks is presented that gives ``longitudinal'' perspectives of the network like radiographs, making the internal structures of the network visible. Impacts of parameter adjustments or structural changes to the network can be visualized via such radiographs. Using WPA, we discover that neural network store and utilize information in a ``holographic'' way, that is, subnetworks encode all training samples in a coherent structure and thus only by investigating the weight pathways can one explore samples stored in the network. Furthermore, with WPA, we reveal fundamental learning modes of a neural network: the linear learning mode and the nonlinear learning mode. The former extracts linearly separable features while the latter extracts linearly inseparable features. The hidden-layer neurons self-organize into different classes for establishing learning modes and for reaching the training goal. The finding of learning modes provides us the theoretical ground for understanding some of the fundamental problems of machine learning, such as the dynamics of learning process, the role of linear and nonlinear neurons, as well as the role of network width and depth, and thus enables us to address the questions of what to learn, how to learn, and how can learn well.

\end{abstract}

\maketitle

\section{INTRODUCTION}\label{}

 In the last decades, learning algorithms have made remarkable progress on numerous machine learning tasks and dramatically improved the state-of-the-art in many practical areas~\cite{RN43,RevModPhys.91.045002}. In addition to practical applications, theoretical studies striving to understand the mechanism of learning models are receiving increased attention~\cite{RevModPhys.91.045002,doi:10.1073/pnas.1907369117,10.1145/3446776}. A particularly interesting topic is to make the ``black-box'' of learning models transparent and explainable~\cite{doi:10.1073/pnas.1907375117,9369420,olah2018the,Bau_2017_CVPR,phillips2019explanatory}. Researchers have carried out studies from various angles, including mutual information~\cite{PhysRevLett.126.200601}, the hidden manifold model~\cite{PhysRevX.8.031003,PhysRevX.10.041044},
 mean-field theory~\cite{doi:10.1073/pnas.1806579115}, stochastic thermodynamics~\cite{PhysRevLett.118.010601}, renormalization group~\cite{PhysRevLett.121.260601},
 statistical mechanics~\cite{Pan,Yoshida_2019,PhysRevLett.124.248302}, etc.. One way to explore the mechanism is to enlist the aid of simple models, particularly linear neural networks, in which single neurons have a linear input-output transfer function~\cite{6177645,NEURIPS2019_c0c783b5,doi:10.1073/pnas.1820226116,RevModPhys.91.045002,PhysRevX.11.031059,PhysRevE.104.034126,PhysRevE.105.064118}. The studies of linear models have achieved great success. Some important network properties are exactly formulated~\cite{PhysRevX.11.031059}, which allow one to gain knowledge on important topics such as network features that give it the ability to generalize despite overparametrization, the role of depth and width of the network, as well as the size of the training set, etc.. However, these approaches treat the network as a whole by deriving analytic equations involving not only weights but also the sample inputs, leaving the network itself a ``black-box'' still. Furthermore, despite sharing important aspects of nonlinear networks, linear neural networks cannot surpass the computational power of a single-layer linear perceptron~\cite{6177645}. Indeed, it is well-known that the success of machine learning in applications is mainly due to the introduction of nonlinear neuron transfer functions. Therefore, developing analysis methods applicable to both linear and nonlinear neural networks and elucidating the role of nonlinearity are still important tasks to pursue.
 \begin{figure}[htp]
  \centering
  \includegraphics[width=1\linewidth]{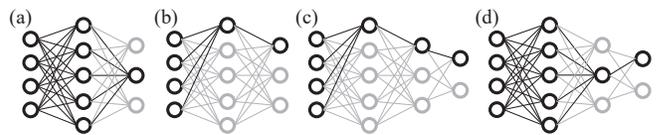}
  \caption{\label{fig:1}Subnetworks of weight pathways, (a) a subnetwork of an output neuron, (b)-(d) subnetworks of hidden-layer neurons. Connections within a subnetwork are depicted by dark lines, while nodes of a subnetwork are represented by dark circles.}
 \end{figure}
 We present a framework based on the concept of weight pathways to study neural networks; we call it weight pathway analysis (WPA). Different from previous layer by layer analyzing approaches, we establish a ``longitudinal'' radiographic view of the network. In more details, considering a neural network shown in Fig.~\ref{fig:1}, we see that there are pathways between input neurons and output neurons that are connected by weighted edges. We call a path connecting an input neuron and an output neuron via these weighted edges a weight pathway and the product of the weights along the path the weight product of the weight pathway. In the case of a monotonic neuron transport function (which is the common case for machine learning; the idea can be extended to more complex neuron transfer functions), the weight product of a weight pathway has a monotonic relation to the neural network outputs. The weight pathway view point gives us ability to investigate direct relations between inputs and outputs of a network.

 With hidden-layer or output-layer neurons as nodes, we decompose a neural network into subnetworks of weight pathways, as examples shown by Fig.~\ref{fig:1}. Then, we integrate all of the weight pathways connecting an input neuron and the output neuron (i.e., sum up the weight products of these weight pathways) and define the value as the penetration coefficient from the input neuron to the output neuron. The penetration coefficients of all of the input neurons to the output neuron of a subnetwork are used to construct a characteristic map of the subnetwork, which characterizes the enhancement or suppression effect of the subnetwork on an input sample's ability to excite that one particular output neuron.

 We visualize the penetration coefficients of a subnetworks and call it the radiograph of the subnetwork. For two-dimensional image samples, the penetration coefficients and the image pixels have a one-to-one correspondence, by which one can infer the enhancement or suppression effect of the subnetwork on each pixel. The radiograph gives a ``longitudinal'' penetrating view of the neural network, making the ``black-box'' less opaque and more interpretable, as the influence of parameter adjustments or structural changes to the network can be visualized via such radiographs. Furthermore, penetrating radiographs enable us to gain important insights on the learning dynamics of neural networks. For example, we discover that subnetworks are ``holographic'', that is, every training sample is encoded in every subnetwork instead of some training samples encode in some subnetwork or neurons. Consequently, we can explore the samples stored in the network by investigating the network weights alone. This discovery provides important insights on how the information is stored in a neural network and how new objects are identified. Our visualization method is different from the famous ``mask'' method~\cite{phillips2019explanatory}, which needs backtracking calculations with the input of samples. It is also different from the map obtained by the backpropagating kernel renormalization for a linear neural network~\cite{PhysRevX.11.031059}, which also involves both the input samples and the weights of the network.

 With the help of penetrating radiographs, we show that hidden-layer neurons self-organize into different classes in order to reduce the cost function. More essentially, we reveal that the network uses a learning mode, we call linear learning mode, for extracting linearly separable features, while uses another learning mode, we call nonlinear learning mode, to extract linearly inseparable features. It is found that linearly separable features can be extracted by single neurons, while linearly inseparable features can only be extracted via cooperation of multiple neurons using the nonlinearity characteristics of their transfer function. As a result, neural networks will always try to extract linear features with linear learning mode first. This explains an important previous finding, that is, the earlier learning stage of a deep neural networks is equivalent to a linear classifier~\cite{NEURIPS2019_b432f34c,NEURIPS2020_c6dfc6b7}. Our study further reveals that the nonlinear learning mode is initiated when linear features are insufficient to further minimize the cost function. Therefore, how to maximize the extraction of linearly separable features and linearly inseparable features is the key to optimize the performance of a neural network. We will show that increasing the width or the depth of a neural network helps this optimization process. This finding provides a theoretical ground for the practice of optimizing a neural network via increasing its width or its depth.

 This paper is arranged as the following. Section II introduces our model and algorithm, where image classification neural networks are employed as illustration examples. Section III introduces core concepts of WPA, including weight pathway, characterizing map, and radiograph for visualizing characterizing maps. In the first part of the Section IV we demonstrates the basic principle of WPA, using toy samples and three-layer neural networks. The linear and nonlinear learning modes, as well as the self-organized differentiation of hidden-layer neurons are shown in this part. The second part of this section applies WPA to the learning of the MNIST set of the handwritten digits, and explains why increasing the width and depth of a neural network can help maximizing the extraction of linearly separable and linearly inseparable features. The final section (V) gives conclusions and discussions.

\section{MODEL and METHOD}\label{sec:2}

 In this paper we consider a multi-layer neural network given by following equations,
 \begin{equation}\label{}
  \begin{aligned}
   \centering
   &x^{(l)}_{i_l} = f(kh^{(l)}_{i_l}) \\
   &h^{(l)}_{i_l} = \sum^{N_{l-1}}_{i_{l-1}} w^{(l)}_{i_l i_{i-1}} x^{(l-1)}_{i_{l-1}},
  \end{aligned}
 \end{equation}
 where $x^{(l)}_{i_l}$ and $h^{(l)}_{i_l}$ are the output and the local field of the $i_l$th neuron in the $l$th layer respectively, $w^{(l)}_{i_l i_{l-1}}$ represents the weight that connects the $i_{l-1}$th neuron in the $(l-1)$th layer to the $i_l$th neuron in the $l$th layer, and $f(\cdot)$ is the neuron transfer function, $N_{l-1}$ is the number of the neurons in the $(l-1)$th layer. Note that we do not involve the neuron bias into the model for sake of the simplicity. Extending the analysis to cases involving biases is straightforward.

 Assuming that the training set is consist of $P$ samples $\{ (\mathbf{x}(\mu), \mathbf{y}(\mu)), \mu=1,\cdots,P \}$, where $\mathbf{x}(\mu)$ is the input vector of the $\mu$th sample with components $x^{(0)}_{i_0}(\mu) (i_0=1,\cdots,N_0)$, $\mathbf{y}(\mu)$ defines its expected state in the output layer. Here we set $y_{i_L}(\mu)=1$ if this sample belongs to the $i_L$th class (we call this neuron the label neuron of this class), and $y_{i_L}(\mu)=-1$ otherwise. For the sake of simplicity, we always apply the linear neuron transfer function to the output layer neurons, i.e., set $x^{(L)}_{i_L}=h^{(L)}_{i_L}$, while applying $f(k h)=k h$ and $f(k h)={\rm tanh}(kh)$ to the hidden-layer neurons in linear neural networks (LNNs) and nonlinear neural networks (NNNs) respectively. We then define the cost function as
 \begin{equation}\label{eq:2}
  S(d) = \frac{1}{P N_L} \sum^P_{\mu=1} \sum^{N_L}_{i_L=1} ( x^{(L)}_{i_L}(\mu) y_{i_L}(\mu)-d )^2,
 \end{equation}
 where $d$ is a parameter that controls the gap between the output of the label neuron and that of the other neurons in the output layer. This cost function is equivalent to the Margin commonly used by support vector machines. Minimizing the cost function to zero gives
 \begin{equation}\label{}
  x^{(L)}_{i_L}(\mu) y_{i_L}(\mu)=d
 \end{equation}
 for all of the samples. We call this condition the goal of training. But as long as $x^{(L)}_{i_L}(\mu)$ is the largest on the label neuron, or equally
 \begin{equation}\label{}
 x^{(L)}_{i_L}(\mu) y_{i_L}(\mu) > 0
 \end{equation}
 for all of the samples, the sample is correctly classified. We call this condition the goal of classification.

 The WPA approach is applicable to all learning algorithms in principle. However, in order to clearly demonstrate how to control the transition from linear learning mode to nonlinear learning mode, we use a simple gradient-free algorithm, namely the Monte Carlo (MC) algorithm~\cite{PhysRevE.70.066137,RN44}. The algorithm is quite simple: select a weight randomly with equal probability, then change it randomly to a new value; accept the change if it reduces the cost-function and discard it otherwise. The operation is repeated until the minimization of the cost function is achieved. Because each update is judged by all samples, the cost-function will be reduced monotonously. Besides, since the changed neurons induced by the mutation involves only those connected by weight paths involving the mutated weight, the computation time is acceptable for neural networks with few hidden layers~\cite{zhao2017general,RN44}.

 Since there is no need of differentiation and back-propagation in this algorithm, we can limit the range of weights, for example, by setting $|w^{(l)}_{i_l i_{l-1}}| \le 1$. Then, with an appropriate transfer function coefficient $k$, we can control the input-output sensitivity of the network to avoid the overlearning. This setting undertakes the weight regularization. Due to the restriction of the norm of the weights, linearly separable feature of a sample cannot be amplified infinitely, and thus the cross-over transition from the linear to nonlinear learning modes can be clearly shown by increasing the parameter $d$ (this is the reason we adopt the cost function with $d$ to be a control parameter). Furthermore, with the restriction on the input-output sensitivity of the network, the width of the network can be greatly increased without overlearning. Therefore, although its training speed is not as fast as other traditional methods, the MC algorithm has a better ability to perform the training when imposing complex restrictions on network parameters and architecture, and thus is helpful for exploring the mechanism of the machine learning from a wider perspective.

 \section{Weight pathways and visualization of penetration coefficients}\label{}

 We call weight edges $w^{(1)}_{i_1 i_0} \rightarrow w^{(2)}_{i_2 i_1} \rightarrow w^{(3)}_{i_3 i_2} \rightarrow \cdots \rightarrow w^{(L)}_{i_L i_{L-1}}$ that connects the $i_0$th neuron in the input layer to the $i_L$th neuron in the output layer a weight pathway.
 We use the product of the weights $w^{(1)}_{i_1 i_0} w^{(2)}_{i_2 i_1} w^{(3)}_{i_3 i_2} \cdots w^{(L)}_{i_L i_{L-1}}$ to characterize a weight pathway and call it the weight product.
 A weight pathway is called a positive pathway if its weight product is positive, otherwise a negative pathway.
 We divide a neural network into subnetworks. Each subnetwork connects all of the input neurons to an output neuron, as illustrated in Fig.~\ref{fig:1}. We then define

 \begin{equation}\label{}
  \begin{aligned}
   \centering
   &c^{(i_l, l)}_{i_0 i_L} = \\
   &\sum_{i_1, i_2, \cdots, i_{l-1}}( w^{(1)}_{i_1 i_0} w^{(2)}_{i_2 i_1} w^{(3)}_{i_3 i_2} \cdots w^{(l)}_{i_l i_{l-1}}) w^{(l+1)}_{i_{l+1} i_l} \cdots w^{(L)}_{i_L i_{L-1}},
  \end{aligned}
 \end{equation}
 as the penetration coefficient of a subnetwork from the $i_0$th input neuron to the $i_L$th output neuron, where the summation is over all of the weight pathways that come out from the $i_0$th input neuron, go through the $i_l$th neuron (we call it the node neuron of the subnetwork) in the $l$th layer, and end with the $i_L$th output neuron. Note that the summation covers all hidden-layer neurons between the $i_0$th input neuron and the $i_l$th neuron (the node neuron) in the $l$th layer. Depending on the selection of $l$ or the node neuron of the subnetwork, we can have summation over weight pathways that pass through all the hidden-layer neurons $(l=L)$ as in Fig.~\ref{fig:1}(a), over a single weight pathway from an input to an output $(l=1)$ as in Fig.~\ref{fig:1}(b) and \ref{fig:1}(c), or over weight pathways through a portion of the hidden neurons $(1 < l < L)$ as in Fig.~\ref{fig:1}(d).

 The most important feature of a weight pathway is that it connects directly the input neurons and the output neurons of a neural network. In the case of monotonic neuron transport functions, the weight products of the weight pathways monotonously determine the direction of the evolution of the local field of an output neuron that shared by the weight pathways, and thus determines the evolution of the cost function. As a result, when we input the $\mu$th sample, the contribution of its $i_0$th component $x^{(0)}_{i_0} (\mu)$ to the local field $h^{(L)}_{i_L} (\mu)$ of the $i_L$th neuron in the output layer is proportional to $c^{(i_l, l)}_{i_0 i_L} x^{(0)}_{i_0} (\mu)$, and the map
 \begin{equation}\label{eq:}
  H^{(i_l, l)}_{i_L} (\mu) = \sum^{N_0}_{i_0=1} c^{(i_l, l)}_{i_0 i_L} x^{(0)}_{i_0}(\mu),
 \end{equation}
 characterizes the contribution of the whole subnetwork to the local field $h^{(L)}_{i_L} (\mu)$. Here, $i_l$ is the index of the node neuron in layer $l$. Note that in the particular case of $l=L$ (as in the case of Fig.~\ref{fig:1}(a)), for a LNN, $H^{(i_L, L)}_{i_L} (\mu) = h^{(L)}_{i_L} (\mu)$, and for a NNN $H^{(i_L, L)}_{i_L} (\mu) \propto h^{(L)}_{i_L} (\mu)$.
 Therefore, penetration coefficients characterize the influence of each component of an input vector on an output neuron by a specific subnetwork of weight pathways; it helps us to infer whether such an input component's influence over that output neuron is positive or negative by the subnetwork. By decomposing a neural network into subnetworks, we can investigate how each part of the neural network works individually and cooperatively to achieve the goal of classification and the goal of training.

 Suppose samples are given by images (with single grey channel) that consist of $M \times M$ bitmap pixels, and the representation vector of the $\mu$th sample is coded as $(x^{(0)}_{i_0}, i_0=M\alpha + \beta, \alpha=1,\cdots,M, \beta=1,\cdots,M)$.
 By plotting a two-dimensional heat map as the following
 \begin{equation}\label{eq:}
  l^{(i_l, l)}_{\alpha \beta} (i_L) = c^{(i_l, l)}_{i_0 i_L},
 \end{equation}
 we have a visualization of the penetration coefficients, where $i_l$ is the index of the node neuron in layer $l$ and $i_L$ is the index of the output neuron in layer $L$. Note that penetration coefficient $c^{(i_l, l)}_{i_0 i_L}$ can either be positive or negative, corresponding to positive pathway dominant or negative pathway dominant respectively, and the corresponding heat map $l^{(i_l, l)}_{\alpha \beta} (i_L)$ pixels are positively displayed (in shades of red pseudo colors) or negatively displayed (in shades of blue pseudo colors) respectively. The resulting heat map images will have patterns of positively or negatively displayed regions, corresponding to regions of input pixels that would be connected to the output neuron with positive or negative penetration coefficients. With these patterns, one can infer the enhancement or suppression effect to a given sample image at the pixel level following the characteristic map. The visualization can be considered a view of the internal coherence structure of a subnetwork. We thus call heat map $l^{(i_l, l)}_{\alpha \beta} (i_L)$ the radiograph of the subnetwork that connects the input neurons to the $i_L$th output neuron through the $i_l$th node neuron in the $l$th layer. We call patterns in the radiograph modes of the subnetwork or the node neuron.

\section{RESULTS}\label{}

\subsection{WPA approach to Toy samples}\label{}

 We illustrate the basic principles of the WPA approach with three training sets of toy samples. The input samples are $100 \times 100$ bitmaps, and they all contain an identical circle but at different positions. The pixels inside the circle (the face zone) are assigned a value of $e=1, 2,$ and $3$ for the first, second and third samples in all three training sets, while the pixels outside of the circle (the ground zone) are always assigned a value of $-1$.
 The three samples of the first training set have no overlap between the face regions, while those of the second training set have completely overlapped face regions, and finally those of the third set have partially overlapped face regions. Each sample can be vectorized as a $N = 100 \times 100 = 10^4$ dimensional vector. We train a \text{$10^4$}--200--3 neural network to perform the task of classification; the cost function is defined by Eq.~(\ref{eq:2}).
 The three samples in each training set correspond to three classes respectively; the first to the third output neurons are the label neurons of the first to the third class, respectively. The achievement of classification means $h^{(2)}_{i_2} (\mu)$ is the largest if $\mu=i_2$, while the achievement of the training goal means $h^{(2)}_{i_2} (\mu) y_{i_2} (\mu) = d$ for every $\mu$ and $i_2$. It will be seen that the second sample of the second set contains only linearly inseparable features and thus is linearly inseparable, while all the other samples contains linearly separable features and are linearly separable, with the second sample of the third set having both linearly separable and inseparable features.

 \begin{figure}[htp]
  \centering
  \includegraphics[width=1\linewidth]{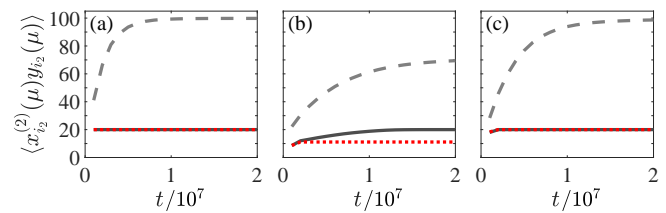}
  \caption{\label{fig:2}The average local field of output neurons $\langle x^{(2)}_{i_2}(\mu) y_{i_2}(\mu)  \rangle$ as a function of training time (MC steps). (a)-(c) correspond to the three training sets, and the dotted lines(red), solid lines(black), and dashed lines(gray) in each plot correspond to LNN20, NNN20, and NNN200 neural networks, respectively.}
 \end{figure}

 As a reference, we first train these three training sets with linear neurons with a transfer function $f(kh)=0.002h$ and with $d=20$. We shorthand this neural network as LNN20. The local fields of output neurons are averaged over the three output neurons and over the three samples as $\langle x^{(2)}_{i_2} (\mu) y_{i_2} (\mu) \rangle$. The time evolution of $\langle x^{(2)}_{i_2} (\mu) y_{i_2} (\mu) \rangle$ for the first to the third training sets are shown as red-dotted lines in Fig.~\ref{fig:2}(a) to \ref{fig:2}(c), respectively. We see that the goal of training is reached as $\langle x^{(2)}_{i_2} (\mu) y_{i_2} (\mu) \rangle \approx 20$ for both the first and the third sets, while $\langle x^{(2)}_{i_2} (\mu) y_{i_2} (\mu) \rangle \approx 11$ for the second set. Averaging only over the three samples of the second training set, we found that $\langle x^{(2)}_{1} (\mu) y_{1} (\mu) \rangle \approx \langle x^{(2)}_{3} (\mu) y_{3} (\mu) \rangle \approx 20$ for the first and the third output neurons, but $\langle x^{(2)}_{2} (\mu) y_{2} (\mu) \rangle \approx 0$ for the second output neuron. These results imply that the goal of classification is reached for the first and the third samples, while it fails for the second sample.

 We then train these training sets with nonlinear transfer function $f(kh)={\rm tanh}(0.002h)$ and with $d=20$. We shorthand this neural network as NNN20. The average local fields of the output neurons $\langle x^{(2)}_{i_2} (\mu) y_{i_2} (\mu) \rangle$ for the first to the third training samples are shown as black-solid lines in Fig.~\ref{fig:2}(a) to \ref{fig:2}(c), respectively. It can be seen that the goal of training is completely fufilled for all the three training sets.

 When $d$ is increased to $200$ for the nonlinear neural network (shorthanded as NNN200), the goal of classification can still be achieved for all training sets, but the goal of training is not fully achieved by any of the training sets, as shown by the gray-dashed lines in Fig.~\ref{fig:2}(a) to \ref{fig:2}(c). For the first and third sets, $\langle x^{(2)}_{i_2} (\mu) y_{i_2} (\mu) \rangle \approx 100$. In details $\langle x^{(2)}_{1} (\mu) y_{1} (\mu) \rangle \approx \langle x^{(2)}_{2} (\mu) y_{2} (\mu) \rangle \approx \langle x^{(2)}_{3} (\mu) y_{3} (\mu) \rangle \approx 100$ for these two sets. For the second set, $\langle x^{(2)}_{i_2} (\mu) y_{i_2} (\mu) \rangle \approx 70$, and  $x^{(2)}_{i_2} (\mu) y_{i_2} (\mu)$ has different value for different $\mu$ and ${i_2}$.

 \begin{figure}[htp]
  \centering
  \includegraphics[width=1\linewidth]{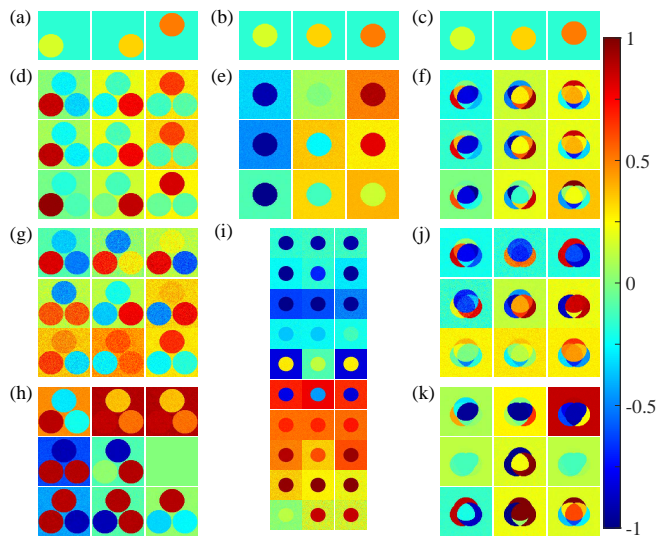}
  \caption{\label{fig:3}Visualization of penetrating coefficients of subnetworks. (a)-(c) show the first to the third training sets. In all other plots, (d)-(k), the first to the third column correspond to the first to the third output neuron, while the rows correspond to different quantities. In more detail, (d)-(f) are radiographs of penetration coefficients of subnetworks of the output neurons trained by the three training sets, respectively. The first to the third rows correspond to LNN20, NNN20 and NNN200 networks, respectively. (g)-(h) are self-organization radiographs of subnetworks of the first training set classed by LNI for NNN20 and NNN200 networks, respectively. The first to third rows correspond to the LNI of the first to the third samples, i.e., $\Pi(\mu|i_2)$ with $\mu=1,2,3$. (i) are self-organization radiographs of subnetworks of all ten hidden-layer neuron classes of the second training set listed in Table I. The ten rows correspond to ten hidden-layer neuron classes. (j)-(k) are same as (g)-(h) but for the third training set.}

 \end{figure}

\subsubsection{Holographic structure}\label{}

 The radiograph of a subnetwork reveals the pattern of penetration coefficients of its characteristic map, which can be used to infer on how the subnetwork works.
 Figure~\ref{fig:3}(a)-\ref{fig:3}(c) show the three samples of the three training sets, respectively, where circles with value $e=1, 2,$ and $3$ are shown in yellow, light brown and dark brown colors.
 After the local fields of the output neurons no longer vary with MC adaptations, we stop the training and show radiographs of subnetworks of the three output neurons $l^{(1, 2)}_{\alpha \beta} (1), l^{(2, 2)}_{\alpha \beta} (2),$ and $l^{(3, 2)}_{\alpha \beta} (3)$ for LNN20, NNN20, and NNN200 in Fig.~\ref{fig:3}(d)-\ref{fig:3}(f), where the columns correspond to the three output neurons and the rows correspond to the three types of neural networks. For better visualization, we calculate an ensemble average of penetration coefficients of $16$ replicas of the trained neural networks for each training set.
 We see that each visualization image clearly shows the patterns of all the samples in the training set, indicating that each subnetwork ``holographically'' encodes all samples of a training set and thus only by investigating weight pathways one can reveal the samples been stored in the network.
 Some zones are positively displayed while others negatively displayed, implying that subnetworks of weight pathways establish coherent structures for enhancement or suppression of different features.
 The ``holographic'' structure reveals how a neural network stores information, and indicates that classifications are achieved through the interaction between the input samples and the ``holographic'' structure. In the following, we will examine how the ``holographic'' structures are formed and in what a way they work.

 \begin{figure*}[htp]
  \centering
  \includegraphics[width=1\linewidth]{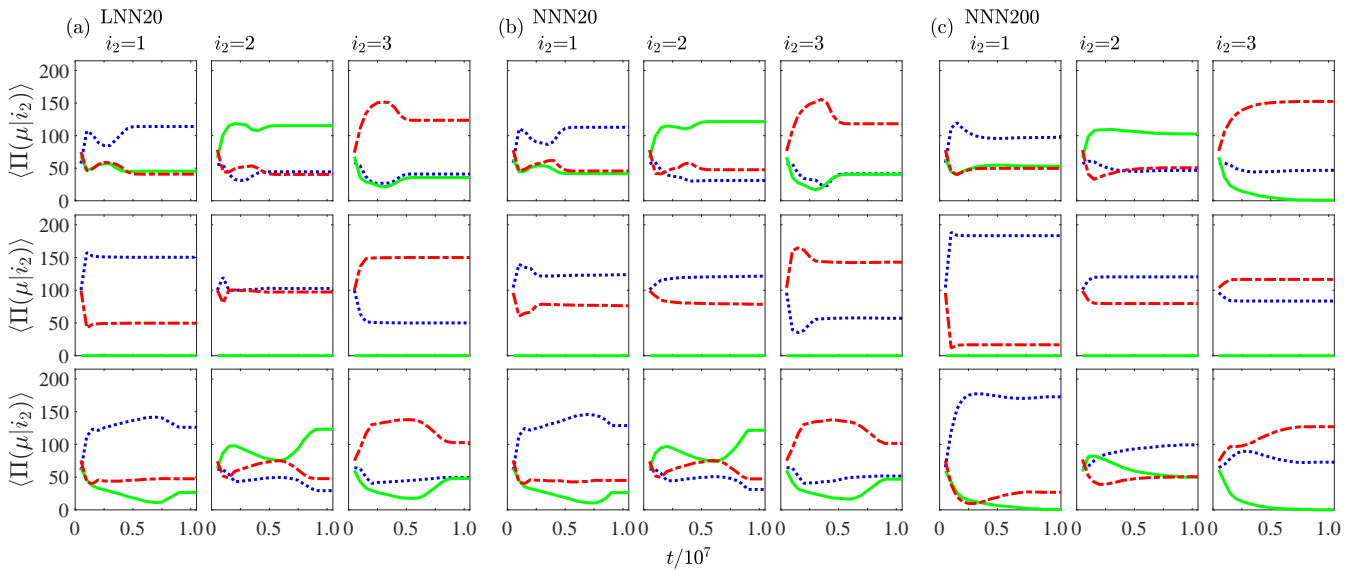}
  \caption{\label{fig:4} The LNI as a function of training time (MC steps). (a)-(c) are for the LNN20, NNN20, and NNN200 neural network, respectively. The first to third rows are for the first to three training sets, and $i_2 = 1,2,3 $ represent the results of the first to three output neurons. In each plot, three colored lines (blue-dotted, green-solid, and red-dashed-dotted) represent the results for the first to the third samples.}
 \end{figure*}

\subsubsection{The extraction of linearly separable feature and the differentiation of hidden-layer neurons.}\label{}

 We study the neural network trained by the first training set in detail in this section. For this set, the characteristic map can be decomposed into four parts, $H^{(i_2, 2)}_{i_2} (\mu) = c({\rm G})x({\rm G}) + c({\rm F_1})x({\rm F_1}) + c({\rm F_2})x({\rm F_2}) + c({\rm F_3})x({\rm F_3})$, which respectively represent the contributions of the common ground zone (zone ${\rm G}$), the face zone of the first sample (zone ${\rm F_1}$), the face zone of the second sample (zone ${\rm F_2}$), and the face zone of the third sample (zone ${\rm F_3}$).
 Here, coefficients $c({\rm G}), c({\rm F_1}), c({\rm F_2}),$ and $c({\rm F_3})$ represent the penetration coefficients, and  $x({\rm G}), x({\rm F_1}), x({\rm F_2}),$ and $x({\rm F_3})$ represent the pixels in the zones ${\rm G}, {\rm F_1}, {\rm F_2}$ and ${\rm F_3}$, respectively. As shown by radiographs of $l^{(1, 2)}_{\alpha \beta} (1), l^{(2, 2)}_{\alpha \beta} (2),$ and $l^{(3, 2)}_{\alpha \beta} (3)$ for LNN20, NNN20, and NNN200 in Fig.~\ref{fig:3}(d), the penetration coefficients within each zone have almost identical values, and the enhancement or suppression effect on output neurons can be identified by the signs of them. Hence, the signs of coefficients $c({\rm G}), c({\rm F_1}), c({\rm F_2}),$ and $c({\rm F_3})$ are employed to tag the modes of the subnetworks or the node neurons. It can be easily seen from Fig.3(d) that the modes of the subnetworks corresponding the three output neurons are $(-,+,-,-), (+,-,+,-),$ and $(+,-,-,+)$, respectively, and remain the same for all three rows, meaning that the learning dynamics for either a LNN, a small parameter $d$ NNN, or a large parameter $d$ NNN are almost same for this training set.
 Given the input pixel values in the four zones $x({\rm G}), x({\rm F_1}), x({\rm F_2})$, and $x({\rm F_3})$ of the first to the third samples as $(-1,1,-1,-1), (-1,-1,2,-1),$ and $(-1,-1,-1,3)$ respectively, we can infer the polarities of the contributions from each zone of an input sample to each output neuron, i.e., we obtain the signs of the contribution of each zone to characteristic map $H^{(i_2, 2)}_{i_2} (\mu)$.
 For example, the first output neuron has $H^{(1,2)}_1 (1) \sim (+,+,+,+), H^{(1,2)}_1 (2) \sim (+,-,-,+),$ and $H^{(1,2)}_1 (3) \sim (+,-,+,-)$, from the first, the second, and the third samples, respectively. We see that the contribution from the ${\rm F_1}$ zone to $H^{(1, 2)}_1 (1)$ is positive, while that to $H^{(1, 2)}_1 (2)$ and $H^{(1, 2)}_1 (3)$ are both negative. So as long as $c({\rm F_1})$ is large enough, the signs of $H^{(1, 2)}_1 (\mu)$ would be identical to that of $c({\rm F_1})x({\rm F_1})$, and it means that the condition of $H^{(1, 2)}_1 (1) > 0, H^{(1, 2)}_1 (2) < 0,$ and $H^{(1, 2)}_1 (3) < 0$ or the goal of classification can always be reached for the first output neuron. Similar argument can be applied to the second and third samples, and one can thus get positive local field for label neurons $(\mu = i_L)$ while keeping all the non-label neurons $(\mu \ne i_L)$ negative. Hence, given the modes of the subnetworks, one can infer how the goal of classification is achieved.

The goal of classification just asks $h^{(2)}_{i_2} (\mu) y_{i_2} (\mu) > 0$; however, the goal of training requires $h^{(2)}_{i_2} (\mu) y_{i_2} (\mu) = d$, which generates additional tasks for the learning dynamics. In the following, we show that the differentiation of hidden-layer neurons is necessary for achieving the goal of training. Since radiographs of the output neurons sum up contributions from all hidden-layer neurons, they cannot be used to explore the differentiation of hidden-layer neurons. One way to capture the self-organization property is to count the number of hidden-layer neurons that give the largest contribution to an output neuron when inputting a specific sample and study the distribution of such counts. We denote the number of the largest-contribution hidden-layer neurons to the $i_2$th output neuron when inputting sample $\mu$ as $\Pi(\mu|i_2)$, and call it largest contribution hidden-layer neuron classification index, or simply the largest neuron index (LNI). To achieve the goal of classification, ideally, all the hidden-lay neurons should maximally contribute to the label neuron and thus appear statistically identical. Deviation from such a behavior is an indication of the occurrence of differentiation of the hidden-layer neurons.

Figure~\ref{fig:4} shows the training time evolution of the LNIs (i.e., $\Pi(\mu|i_2)$) in networks LNN20, NNN20, and NNN200. We see that while the hidden-layer neurons giving the largest contribution to the label neuron do have the largest LNI (i.e., $\Pi(\mu|i_2)$) when $\mu = i_2$, there are still a lot of hidden-layer neurons give their largest contribution to the non-labeled output neurons. This fact reveals the differentiation and self-organization of the hidden-layer neurons.

Self-organization differentiation provides additional degrees of freedom for hidden-layer neurons to achieve the goal of training. To reveal this property, we show self-organization radiographs $l^{(\Pi(\mu|i_2),1)}_{\alpha \beta}$ of network NNN20 in Fig.~\ref{fig:3}(g), and of network NNN200 in Fig.~\ref{fig:3}(h). Here the index $\Pi(\mu|i_2)$ specifies that the radiograph is obtained by only hidden-layer neurons of $\Pi(\mu|i_2)$ class. Self-organization radiographs for the LNN are similar to that of NNN20 and are therefore not shown here. We emphasize again that the radiographs here are produced from the weight pathways alone, without the need of inputting any sample; the samples referenced in the self-organization radiographs are only utilized in their capacity to classify the hidden-layer neurons.

We look at radiographs $l^{(\Pi(\mu|1),1)}_{\alpha \beta}$ of hidden-layer neuron class $\Pi(\mu|1)$ toward the first output neuron in network NNN20 in the first column of Fig.~\ref{fig:3}(g) as an example. The plot of the first row gives the radiograph $l^{(\Pi(1|1),1)}_{\alpha \beta}$ of $\Pi(1|1)$ class neurons. The signs of coefficients $c({\rm G}), c({\rm F_1}), c({\rm F_2})$, and $c({\rm F_3})$, i.e., the mode of these neurons, are $(-,+,-,-)$, which can be seen from the radiograph. The characteristic maps of this neuron class when inputting the first, the second, and the third samples are $H^{(\Pi(1|1),1)}_1 (1) \sim (+,+,+,+), H^{(\Pi(1|1),1)}_1 (2) \sim (+,-,-,+),$ and $H^{(\Pi(1|1),1)}_1 (3) \sim (+,-,+,-)$, respectively. We see that the contribution of the ${\rm F_1}$ zone is positive with the first sample inputting, and is negative with the second or third sample inputting. Consequently, the goal of classification can be achieved as long as $c({\rm F_1})$ is positively large enough. This class of hidden-layer neurons plays the role of reaching the goal of classification.

Similarly, radiograph $l^{(\Pi(2|1),1)}_{\alpha \beta}$ of the $\Pi(2|1)$ neuron class (the second row) has the mode  $(+,+,+,-)$,
giving characteristic maps of this neuron class as $H^{(\Pi(2|1),1)}_1 (1) \sim (-,+,-,+), H^{(\Pi(2|1),1)}_1 (2) \sim (-,-,+,+),$ and $H^{(\Pi(2|1),1)}_1 (3) \sim (-,-,-,-)$. Radiograph $l^{(\Pi(3|1),1)}_{\alpha \beta})$ of $\Pi(3|1)$ neuron class (the third row) shows mode  $(+,+,-,+)$, giving $H^{(\Pi(3|1),1)}_1 (1) \sim (-,+,+,-), H^{(\Pi(3|1),1)}_1 (2) \sim (-,-,-,-),$ and $H^{(\Pi(3|1),1)}_1 (3) \sim (-,-,+,+)$.  Therefore, optimization of the distribution of the three classes can help the network toward the goal of training. For example, increasing the number of $\Pi(1|1)$ class helps evolving $h^{(2)}_1 (1)$ toward $d$ while increasing that of $\Pi(2|1)$ or $\Pi(3|1)$ classes helps advancing $h^{(2)}_1 (3)$ or $h^{(2)}_1 (2)$ toward $-d$. Hidden-layer neurons in the  $\Pi(2|1)$) and  $\Pi(3|1)$) classes play auxiliary role for reaching the goal of training.

We have seen that radiographs of the output neurons are qualitatively similar for either the LNN and the two NNNs.  We see here that radiographs of hidden-layer neurons show certain difference between the NNN20 and NNN200, e.g, the G zones of the radiographs of NNN200 almost all reverse their signs relative to those of the NNN20. The difference is a reflection of the network's effort to reach the goal of training for the much larger parameter $d$. This demonstrates the versatility and power of WPA, which can peek deeper into the functional structure of the network by targeting hidden-layer neurons directly, revealing details of the learning dynamics of the network.

 \begin{figure}[htp]
  \centering
  \includegraphics[width=1\linewidth]{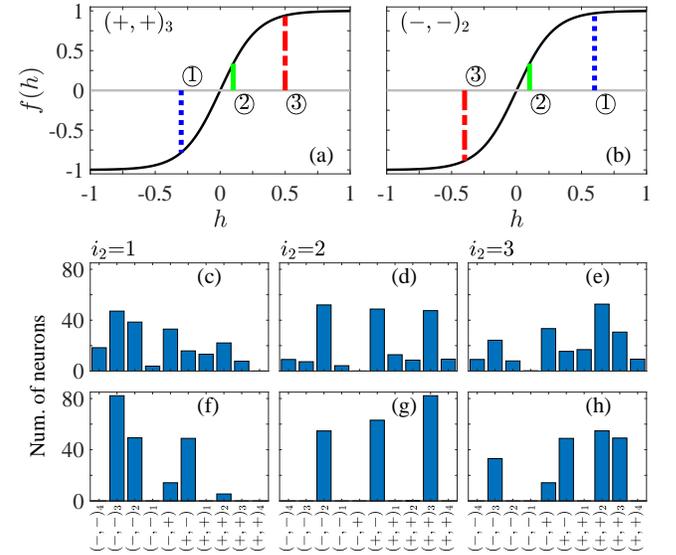}
  \caption{\label{fig:5} (a) and (b) show the formation of the nonlinear learning mode: The modes of $(+,+)_3$ and $(-,-)_2$ of hidden-layer neurons which together forms the nonlinear learning mode that converts the second sample the largest. (c) to (e), and (f) to (h) show the number of hidden-layer neurons of the ten modes (see Table I) for the NNN20 and NNN200,respectively, confirming the differentiation of hidden-layer neurons. The output neuron is indicated by $i_2$}
 \end{figure}

 \begin{table}
  \caption{\label{tab:1}{Modes of hidden-layer neurons for the second training set}}
  \begin{ruledtabular}
  \begin{tabular}{lcr}
  modes & $H^{(i_l, l)}_{i_2} (\mu)$\\
  \hline
  $(-,-)_4$ & $0 > H^{(i, l)}_{i_2} (1) > H^{(i, l)}_{i_2} (2) > H^{(i, l)}_{i_2} (3)$\\
  $(-,-)_3$ & $0 < H^{(i, l)}_{i_2} (1), 0 > H^{(i, l)}_{i_2} (2) > H^{(i, l)}_{i_2} (3)$\\
  $(-,-)_2$ & $H^{(i, l)}_{i_2} (1) > H^{(i, l)}_{i_2} (2) > 0, H^{(i, l)}_{i_2} (3) < 0$\\
  $(-,-)_1$ & $H^{(i, l)}_{i_2} (1) > H^{(i, l)}_{i_2} (2) > H^{(i, l)}_{i_2} (3) > 0$\\
  $(-,+)$ & $0 < H^{(i, l)}_{i_2} (1) < H^{(i, l)}_{i_2} (2) < H^{(i, l)}_{i_2} (3)$\\
  $(+,-)$ & $0 > H^{(i, l)}_{i_2} (1) > H^{(i, l)}_{i_2} (2) > H^{(i, l)}_{i_2} (3)$\\
  $(+,+)_1$ & $0 < H^{(i, l)}_{i_2} (1) < H^{(i, l)}_{i_2} (2) < H^{(i, l)}_{i_2} (3)$\\
  $(+,+)_2$ & $H^{(i, l)}_{i_2} (1) < H^{(i, l)}_{i_2} (2) < 0, H^{(i, l)}_{i_2} (3) > 0$\\
  $(+,+)_3$ & $0 > H^{(i, l)}_{i_2} (1), 0 < H^{(i, l)}_{i_2} (2) < H^{(i, l)}_{i_2} (3)$\\
  $(+,+)_4$ & $0 < H^{(i, l)}_{i_2} (1) < H^{(i, l)}_{i_2} (2) < H^{(i, l)}_{i_2} (3)$\\
  \end{tabular}
  \end{ruledtabular}
 \end{table}

\subsubsection{The extraction of linearly inseparable feature}\label{}

 For the second training set, every sample shares the same face zone $({\rm F})$ and ground zone $({\rm G})$. The characteristic map appears as $H^{(i_l, l)}_{i_2} (\mu) = c({\rm G})x({\rm G})+c({\rm F})x({\rm F})$. Given $x({\rm G})=-1$ and $x({\rm F}) = 1, 2, 3$ for the three samples, and depending on the amplitudes of $c({\rm G})$ and $c({\rm F})$, which can be used to perform a detailed analysis about the possible modes of subnetworks or of the hidden-layer node neurons. In terms of signs of penetration coefficients, hidden-layer neurons have four possible modes, i.e., $(-, -)$, $(+, +)$,  $(+, -)$ and $(-, +)$. We further consider the amplitude of polarizations of coefficients of $(c({\rm G}), c({\rm F}))$ zone, and expand to a total of $10$ hidden-layer modes (a subscript is used for modes with the different combinations of signs of $H^{(i_l, l)}_{i_2} (\mu)$), which are listed in Table~\ref{tab:1}. These results are applicable for both linear and nonlinear neurons if the neuron transport function is monotonous. Among the list modes, neuron modes $(+, -)$ and $(-, -)_j$ can all make the first output neuron the largest for the first sample, and neuron modes of $(-, +)$ and $(+, +)_j$ can all make the third output neuron the largest for the third sample, among which $(-, -)_3$ and $(+, +)_2$ can guarantee the achievement of the goal of classification for these two samples.

 However, we see that there is no mode that can make the second output neuron the largest for the second sample. This is because its face zone is identical to those of the other samples and its pixel value is between that of the first and third ones; therefore, the monotonic transformation cannot make its output largest in any way. In other words, a single neuron or a single subnetwork alone cannot realize the classification of the second sample. These predictions are confirmed by the LNI shown in the second rows of Fig.~\ref{fig:4}, where $\Pi(2|i_2)$ vanishes while $\Pi(1|i_2)$ and $\Pi(3|i_2)$ both have significant presence.

 Obviously, the first and third samples are linearly separable as confirmed by the fact that a LNN can classify them, and the second sample is linearly inseparable since a LNN cannot classify it. Therefore, the fact that the NNN can achieve the goal of classification of the second sample (see Fig.2(b)) implies that it uses combinations of subnetworks with different modes. As shown in Fig.~\ref{fig:5}(a) and \ref{fig:5}(b), a pair of hidden-layer neurons with $(+, +)_3$ and $(-, -)_2$ modes can convert the output of the second sample to be the largest, as long as the outputs of the first and third samples lie on the nonlinear region of transfer function ${\rm tanh} (kh) \approx \pm 1$. In this case, the outputs of the first and third samples cancel approximately, while that of the second sample appears as a large positive quantity since both neurons contribute a positive term. We emphasize that the nonlinearity of the transfer function plays a key role here. With a linear neural transfer function, $f(kh) = kh$, $(+, +)_3$ and $(-, -)_2$ must lead to the combined outputs of the second sample vanishing if one wants to offset the outputs of the first and the third samples.

 For this training set, the realization of the goal of training also requires the differentiation of hidden-layer neurons. For example, the first output neuron can achieve the goal of classification with hidden-layer neurons in the $(-, -)_3$ mode, since it gives $h^{(2)}_2(1)>0$ and $0 > h^{(2)}_2(2) > h^{(2)}_2(3)$. However, the goal of training, $h^{(2)}_2(2) = h^{(2)}_2(3) = -d$, cannot be achieved by only this mode. Therefore, auxiliary hidden neurons with different mode must be utilized. Since there are only ten modes for this training set, we can check the distribution of hidden neurons for all of the possible modes for a more detailed investigation of the differentiation. In Fig.~\ref{fig:5}(c) to 5(e), and \ref{fig:5}(f) to 5(h), we show the number of hidden-layer neurons of the ten modes for the NNN20 and NNN200, respectively. Clearly, hidden-layer neurons are differentiated into different modes.

 We see that only one mode vanishes in NNN20 relating to each output neuron. The existing multiple neuron modes provides sufficient freedom to satisfy the goal of training. This is the reason that the condition $\langle x_{i_2}^{(2)}(\mu)y_{i_2}(\mu) \rangle = 20$ is fully reached for all samples and all output neurons as shown in Fig.2. In NNN200, for the first and the third output neurons, the amount of models can still lead to $\langle x_{i_2}^{(2)}(\mu)y_{i_L}(\mu) \rangle = d$ with $d \sim 100$ as shown in Fig. 2, but cannot reach the goal of $d =200$ because the total number of hidden-layer neurons is not enough. For the second output neuron (Fig.5(g)), we see that only modes $(+,+)_3$, $(-,-)_2$ and $(+,-)$ are remained. They are most essential for the goals of both classification and training. The combination of the first two guarantees the output of the second sample the largest, while lead to $h^{(2)}_2(1)$ and $h^{(2)}_2(3) \approx 0$ (see Fig. 5 (a) and 5(b)). Then, the last one, as can be realized from Table I, decrease the outputs globally toward $h^{(2)}_2(1) = h^{(2)}_2(3)=-d$. It is obvious that neurons with these three modes cannot lead to  $x_{i_2}^{(2)}\mu)y_{i_L}(\mu)$ reaches the same value for different samples, explaining the result of Fig.2 (b) for the NNN200.

 In Fig.~\ref{fig:3}(i), we show radiographs $l^{(C_i, 1)}_{\alpha \beta}$ of subnetworks of the ten modes for the NNN20. We see that they are in well agreement with modes shown in Table~\ref{tab:1}. Note that radiograph $l^{(i_2, 2)}_{\alpha \beta}(i_2)$ of Fig.~\ref{fig:3}(e) is the result of superposition of radiographs of the ten classes. For the first and the third output neurons, since hidden-layer neurons with modes $(-, -)_3$ and $(+, +)_2$ have the largest populations respectively, $l^{(1, 2)}_{\alpha \beta}(1)$ and $l^{(3, 2)}_{\alpha \beta}(3)$ thus show patterns of these two modes respectively. For the second output neuron, hidden-layer neurons with modes of $(+, +)_3$ and $(-, -)_2$ have roughly equal numbers but opposite signs (see Fig.~\ref{fig:5}(d)), and hence patterns resulted from them are almost canceled with each other. The pattern of $l^{(2, 2)}_{\alpha \beta}(2)$ is caused by neurons of mode $(+,-)$, since according to Fig.~\ref{fig:5}(d) this class has a large number of hidden-layer neurons.

 \begin{figure}[htp]
  \centering
  \includegraphics[width=1\linewidth]{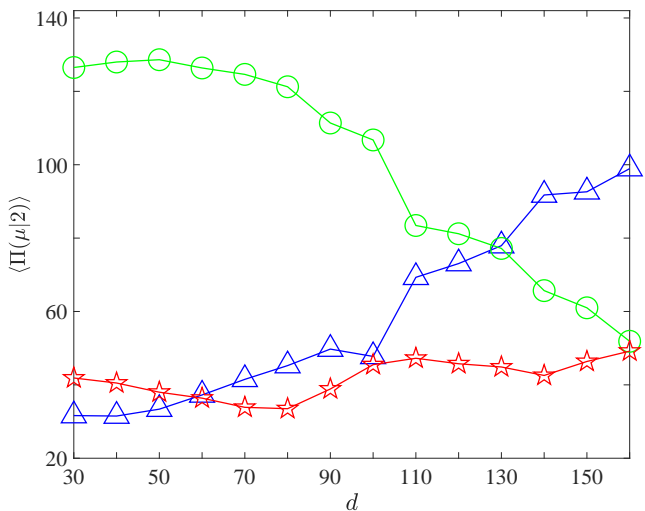}
  \caption{\label{fig:6}The transition from the linear learning mode phase to the linear and nonlinear learning modes mixing phase. The triangles, circles, and stars represent the stationary values of $\Pi(1|2)$, $\Pi(2|2)$ and $\Pi(3|2)$ for a NNN trained by the third training set with increasing $d$.}
 \end{figure}

 Therefore, extracting linearly separable and inseparable features has an essential difference. For the linearly separable feature, neurons can independently achieve the goal of classification. The differentiation of hidden-layer neurons is for the purpose of achieving the goal of training. Examples including all samples in the first training set, as well as the first and the third samples in the second training set. For these samples, a LNN can achieve the goal of classification. We call this mode of learning the linear learning mode. For linearly inseparable features, i.e., the second sample of the second training set, the neural network must use a combination of neurons with different modes and rely on the nonlinearity of the neuron transfer function to make the corresponding output the largest. Linear networks cannot achieve such a goal. We call this mode of learning the nonlinear learning mode.

\subsubsection{The extraction of co-existing linearly separable and inseparable features}\label{}

 In the third training set, the second sample has both linearly separable features (the part of the face zone that do not overlap with face zones of any of the other samples, which shall be called the unit linear features of the sample, and the part of the face zone that only overlap with the face zones of some but not all of the other samples.) and linearly inseparable features (the part of the face zone that overlap with the face zones of all of the other samples). Therefore, it can be used to investigate how the neural network works in the coexistence of linearly separable and inseparable features. The first and third samples have only linearly separable features. For the convenience of discussion below, we denote zones of the unit linear feature of the three samples by ${\rm F^0_1, F^0_2}$, and ${\rm F^0_3}$, respectively.

 The last row of Fig.~\ref{fig:4} shows the training time evolution of the LNI. We see that the LNI of the NNN20 behaves almost exactly as that of the LNN20. Furthermore, Fig.~\ref{fig:3}(f) shows that the radiographs of these two neural networks are similar. Therefore, the NNN20 behaves almost identically to that of the LNN20, indicating that the network invokes only the linear learning mode and extracts only the linearly separable features.

 Radiographs of output neurons can reveal how the linear learning mode works. For example, from $l^{(2, 2)}_{\alpha \beta}(2)$ in Fig.~\ref{fig:3}(f) we see that $c({\rm F^0_2})$ is positive, and $c({\rm F^0_1})$ and $c({\rm F^0_3})$ are negative. Following the characteristic map, the signs of contributions of these three unit linear zones to the outputs should be $H^{(2, 2)}_2(1) \propto c({\rm F^0_1})x({\rm F^0_1})+c({\rm F^0_2})x({\rm G})+c({\rm F^0_3})x({\rm G}) \sim (-, -, +)$, $H^{(2, 2)}_2(2) \propto c({\rm F^0_1})x({\rm G})+c({\rm F^0_2})x({\rm F^0_2})+c({\rm F^0_3})x({\rm G}) \sim (+, +, +)$, and $H^{(2, 2)}_2(3) \propto c({\rm F^0_1})x({\rm G})+c({\rm F^0_2})x({\rm G})+c({\rm F^0_3})x({\rm F^0_3}) \sim (+, -, -)$, respectively. Similar to the first training set, these modes can guarantee the goal of classification, by amplifying sufficiently only the amplitude of positive pathways in ${\rm F^0_2}$ zone for example.

 These facts indicate that applying a NNN does not necessarily mean the extraction of the linearly inseparable features when both linearly separable and linearly inseparable features coexist in a training set. The reason is that the linear learning mode is performed by hidden-layer neurons independently, while the nonlinear learning mode needs to have multiple neurons working cooperatively and thus is more difficult to be manifested. So, if the goal of training can be achieved using only the linearly separable features, the neural network would avoid to activate the nonlinear learning mode. Therefore, one may be at risk of losing the information carried by linear inseparable features in samples with both linearly separable and linearly inseparable features even with a NNN.

 However, using the cost function from Eq. (2) and limiting the norms of the weights, such losses can be avoided. Once the control parameter $d$ exceeds a threshold, weight pathways connecting only the linearly separable zones will not be able to fulfil the goal of training. In this case, the neural network has to initiate the nonlinear learning mode to extract the linearly inseparable features to further drive the local fields towards the goal of training. For the NNN200, comparing to the results of LNN20 and NNN20, we see from Fig.~\ref{fig:3}(f) and Fig.~\ref{fig:4} (c) that there is no qualitative change for the first and the third output neurons for both the radiographs and the LNI's, indicating that the neural network still extracts information with the linear learning mode, since in these cases all the information are the linearly separable feature. However, remarkable changes can be seen in the case of second output neuron. The center zone of $l^{(2,2)}_{\alpha \beta}(2)$ changes to the same pseudo-color as that of the second training set ($l^{(2, 2)}_{\alpha \beta}(2)$ for the two NNNs, see the second and third rows of Fig.~\ref{fig:3}(e)). Fig.~\ref{fig:4}(c) indicates that $\Pi(2|2)$ is no longer the largest, instead, $\Pi(1|2)$ turns to be the largest as also in the case of the second training set.

 These facts indicate that the nonlinear learning mode is initiated for the second sample, i.e., the features of the center zone of this sample, begins to contribute to the local field $h^{(2)}_2(2)$, which can be observed more clearly by studying radiographs of subnetworks of hidden-layer neurons classified by the LNI. Figure~\ref{fig:3} (j) and 3(k) show these images for the NNN20 and NNN200, respectively. For the NNN20, it can be seen that $l^{(\Pi(\mu|i_2),1)}_{\alpha \beta}$ with $\mu=i_2$ (plots along the diagonal line in Fig.3(j)) are qualitatively similar to $l^{(i_2,2)}_{\alpha \beta}(i_2)$ with $i_2=1,2,3$ for both the LNN20 and NNN20 (the first two rows of Fig. 3(f)). This is the result of the linear learning mode neurons keeping dominant, as can be seen from Fig. 4. The $\Pi(2|2)$ class in LNN20 and NNN20 remains the largest population, which indicates that the linear feature of the second sample is utilized dominantly and the patterns of $l^{(2,2)}_{\alpha \beta}(2)$ is dominated by this class.

 For the NNN200, the radiograph of $\Pi(2|2)$ shows non-substantial changes to that of the NNN20, indicating that the main role of the $\Pi(2|2)$ class is still for extracting the linearly separable features of the second sample. This portion of the neurons can still make the second output neuron largest for the second sample. Meanwhile, we see that radiographs of $\Pi(1|2)$ and $\Pi(3|2)$, i.e., $l^{(\Pi(1|2),1)}_{\alpha \beta}$ and $l^{(\Pi(3|2),1)}_{\alpha \beta}$, have changed substantially from that of Fig.~\ref{fig:3}(j), indicating that these two neural classes indeed play the role of modes $(+, +)_3$ and $(-, -)_2$ in Fig.~\ref{fig:5}(a) and \ref{fig:5}(b). The center zone of its radiograph $l^{(\Pi(1|2),1)}_{\alpha \beta}$ is negative, making contribution from the third sample to the second output neuron the most negative or the smallest in value (since $x({\rm F_3})=3$ while $x({\rm F_1})=1$ and $x({\rm F_2})=2$). In addition, from the radiograph we see that the unit linear feature zone ${\rm F^0_2}$ is positive, and thus making the contribution from the second sample larger. As a result, for this neuron class, we have $H^{(\Pi(1|2),1)}_2(1) > H^{(\Pi(1|2),1)}_2(2) > H^{(\Pi(1|2),1)}_2(3)$. Similar analysis can lead to $H^{(\Pi(3|2),1)}_2(1) < H^{(\Pi(3|2),1)}_2(2) < H^{(\Pi(3|2),1)}_2(3)$ for the $\Pi(3|2)$ class. Therefore, when inputting the first or the third sample the outputs from these two neural classes would offset each other; however when inputting the second sample these two neural classes would both contribute to the output positively, and lead to the extraction of the linearly inseparable feature of the second sample.

 Therefore, in the case of the NNN20, the majority of the hidden-layer neurons is employed to extract the linearly separable feature, and the goal of training is achieved by with the help of some auxiliary neurons. In this case with such a small $d$, the excitation of the nonlinear learning mode is not necessary. In the case of NNN200, the goal of training cannot be achieved by learning just the linearly separable features, and the nonlinear learning mode has to be initiated. In Fig.~\ref{fig:6}, we plot $\Pi(1|2), \Pi(2|2)$, and $\Pi(3|2)$ of the NNN as a function of control parameter $d$. Each data point is obtained after the training becomes stationary. It can be seen that $\Pi(2|2)$ begins to decrease around $d = 60$, and $\Pi(1|2)$ begins to increase and becomes dominant after $d = 140$. This plot confirms that the neural network tends to use linear learning mode to extract linear features first, and starts to invoke the nonlinear learning mode only when the linear features are insufficient to approach the goal of training.

 \begin{figure}[htp]
  \centering
  \includegraphics[width=1\linewidth]{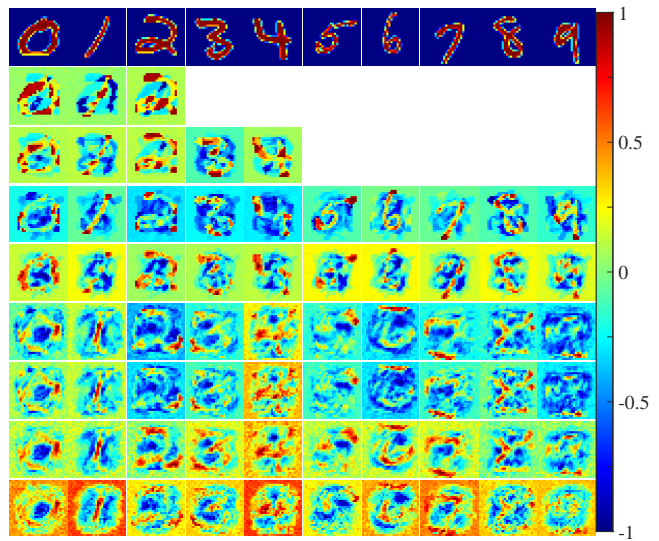}
  \caption{\label{fig:7}The visualization of penetrating coefficients of neural network of MNIST dataset. The 1st row: ten digits picking up from the dataset. The 2nd to the 4th rows are the radiographs of NNNs with $d=30$ trained by the first three, first five and all of the ten digits. The 5th row shows the radiographs of the NNN with $d=75$ trained by the ten digits. The 6th and the 7th rows are the radiographs of the LNN and the NNN of three-layer neural networks trained by the first 600 MNIST samples with the optimum parameters. The 8th and 9th rows are the results of a four-layer and five-layer ones trained by also these samples. The widths of the hidden layers in these neural networks are fixed at $N_1=600$.}
 \end{figure}

\subsection{WPA approach to handwritten-digit samples}\label{}

\subsubsection{Extracting the linearly separable and inseparable features}\label{}

 The first row of Fig.~\ref{fig:7} shows $10$ handwritten digits selected from MNIST. Since each sample is composed of a $28 \times 28$ bitmap, we design a 784--600--P network with $P = 3, 5, 10$ to classify the first $3$, first $5$, and all $10$ digits, with control parameters $d = 30$ and $k= 0.15$. Radiographs of label neurons, i.e., $l^{(i_L, 2)}_{\alpha \beta} (i_L)$ for $i_L = 1, 2, \cdots, P$, are shown in the second to the fourth rows in Fig.~\ref{fig:7}, respectively. The radiographs are obtained after the neural network has been trained to have its cost function less than $0.01$. It can be seen that in the case of only a few classes, the radiographs are similar to those of the toy training sets, i.e., patterns of all the digits in the training set distinctly appear in every radiograph, indicating that the ``holographic'' nature of the neural network remains. With the increase of the number of classes, the pattern of the digit corresponding to a label neuron can still positively recognized, while patterns of other digits gradually become less distinguishable. By studying the progressive trend of radiographs from the cases of $P = 3$ to $P = 5$ and to $P = 10$, one can realize that even for the case of $P = 10$, the patterns of all the digits are indeed still there; they overlap with each other to form the negatively displayed region. In other words, the ``holographic'' structure exists always.

 From the radiographs we can realize that the linear dynamics is similar to that for toy samples.  Note that the zones of each digit correspond to the face zones in the toy training sets.
 Taking the first plot in the second row as an example, we see that the zones of unit linear features (the zones of a digit that do not overlap with other digits) of the digit zero, is positively highlighted. The face zones of digit $1$ and digit $2$ having no overlap with that of the digit zero appear in negative pseudo color, while overlapped part of them are more negative. The face zones of the digit zero that overlap with digits $1$ and $2$ are also positively displayed, but to a lesser extent. Following the characteristic map, these kinds of patterns enhance the output of digit zero ($\mu=1$) on its label neuron to approach $h^{(2)}_1 (1)=d$, while drive the output $h^{(2)}_1 (2)$ of the digit $1$ (($\mu=2$)) and $h^{(2)}_1 (3)$ of digit $2$ ($\mu=3$))toward $-d$. These patterns thus reveal how linearly separable features are extracted and utilized by the linear learning mode.

 The face zones of digits have different pixel amplitudes. Therefore, when several of them overlap in some regions, there must exist the situation that one or more digit whose face is covered between those of others, as as in the case of the second sample of the second toy training set. Such cases may become common when all of the ten digits are involved. Faces been covered represent linearly inseparable features. The fifth row of Fig.~\ref{fig:7} shows the radiographs of a NNN obtained by learning the ten digits with $d = 75$. Comparing to the case of $d = 30$, we see that they are largely similar. In other words, it is difficult to verify whether the nonlinear linear mode is initiated by studying the radiographs alone. The reason is that even if there are linearly inseparable features, they will be fragmented and difficult to identify. This is different from the case of the second sample in the third toy training set, where the linearly inseparable feature zone is large and distinctive. However, we can still see an increasing degree of fragmentation in radiographs of $d = 75$ comparing to the case of $d = 30$, which may be a hint of the appearance of the nonlinear learning mode.

 \begin{figure*}[htp]
  \centering
  \includegraphics[width=1\linewidth]{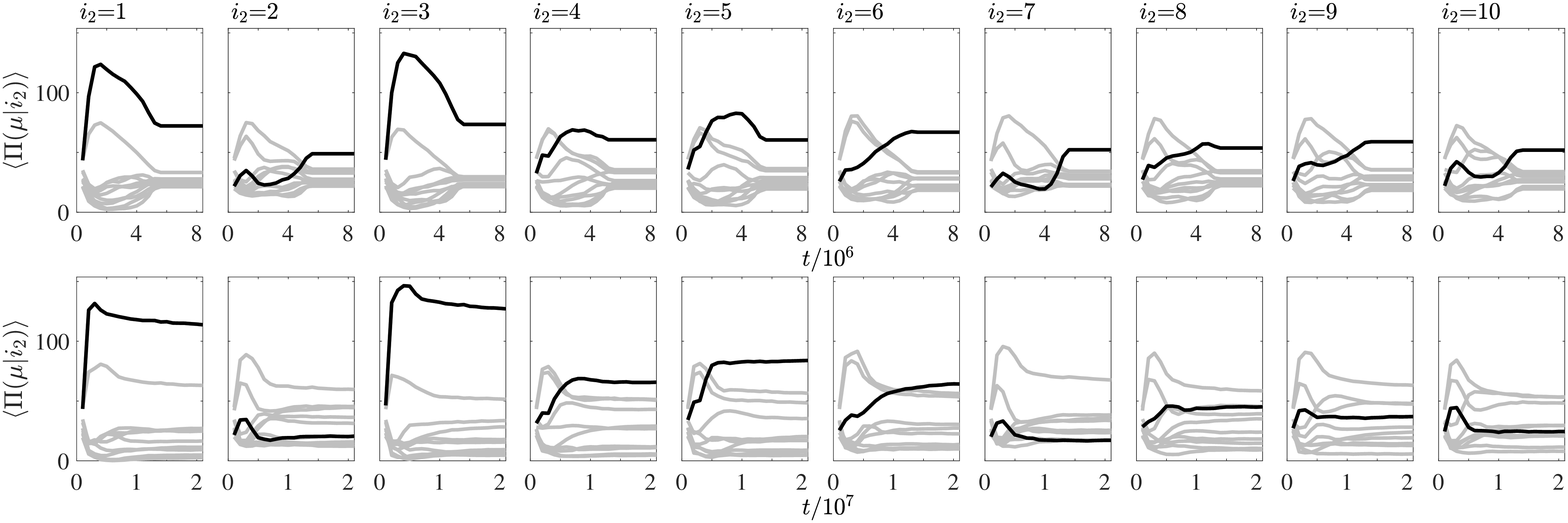}
  \caption{\label{fig:8}The LNI as a function of training time (MC steps) for NNNs with $d=30$ (first row) and $d=75$ (second row). The lines in each panel represent $\Pi(\mu|i_2)$ with $\mu = 1, 2, 3, \cdots, 10$ for specific output neuron $i_2$; the dark line represents $\Pi(\mu|i_2)$ with $\mu = i_2$.}
 \end{figure*}

 Solid evidence of the existence of the nonlinear learning mode can, however, be obtained from LNI. In Fig.~\ref{fig:8}, we show the evolution of the LNI of the NNN trained by the $10$ handwritten digits as a function of time for $d = 30$ and $d = 75$. We see that in the case of $d = 30$,  $\Pi (\mu | i_2)$ is the largest when $\mu = i_2$ for all classes. This fact indicates that the goal of training is mainly achieved by the single class of hidden-layer neurons which can turn the output of the sample to be the largest in its label neuron, and thus the learning process is dominated by the linear learning mode. In the case of $d = 75$, however, the LNI of five digits no longer meet this condition. For example, for $i_2=2$, i.e., the second output neuron, the $\Pi (2|2)$ appears quite low while $\Pi (3|2)$ becomes the largest. To guarantee the digit $1$ instead of $2$ to be the largest to the second output neuron, the positive output of  $\Pi (3|2)$ neuron class must be offset by other neuron classes having negative outputs with the input of digit $2$, in the way similar to the case of Fig.5(a) and 5(b). The training goal is thus partially achieved by the combination of different neuron classes, indicating that the nonlinear learning mode has been initiated. This result confirms that using the cost function with a large $d$ is an effective strategy to excite the nonlinear learning mode.

However, with a large number of samples (and thus $\mu$ has a large freedom) the LNI approach is no longer effective to characterize the excitation of the nonlinear learning mode. In this case, we have to detect the effect of the nonlinear learning mode by checking the accuracy rate that can be reached by the LNN and the NNN. The benefit of the extraction of linearly inseparable feature can be revealed by the improvement of the test accuracy reached by the NNN. To demonstrate this effect, we train a LNN and a NNN of three layers with $N_1=600$ using the first $600$ samples from MNIST. We have checked that the Optimal control parameters are $d=180$ and $k=0.013$ for the LNN and $d=70$ and $k=0.15$ for the NNN. We have checked that the training accuracies for both networks have reached $100$\%. This fact indicates that each training sample contains sufficient linearly separable features for achieving the goal of classification. Figure 9(a) shows the evolution of the accuracy on the supplied test set of MNIST with the training time for these two neural networks. We see that the test accuracy of the nonlinear network is noticeably higher than that of the linear network. This facts indicates that the training samples also contain linearly inseparable features that only a NNN is able to learn and such learnings help it to achieve a noticeable higher accuracy on the test set.

The $6$th and $7$th rows of Fig.~\ref{fig:7} show radiographs of the ten output neurons for the LNN and the NNN at the optimal test accuracies, respectively. We see that patterns of digits emerge also in the radiographs. The positively highlighted parts of a digit represent the zones of common linearly separable features of a class of samples. Again, the radiographs of the LNN and the NNN show less distinctive feature as those in the $4$th and $5$th rows of Fig.~\ref{fig:7}. As revealed by Fig.9(a), the improvement on the test accuracy by the NNN is only about $8$\%. Therefore, in the case of the MNIST data set, the linear separable feature is the dominant feature, so radiographs of MNIST are dominated by linearly separable features, and thus no obvious difference can be discerned between the radiographs with or without the nonlinear learning mode. Nevertheless, the emerged patterns indicate that our visualization approach works also for neural networks trained by large number of samples.

 \begin{figure}[htp]
  \centering
  \includegraphics[width=1\linewidth]{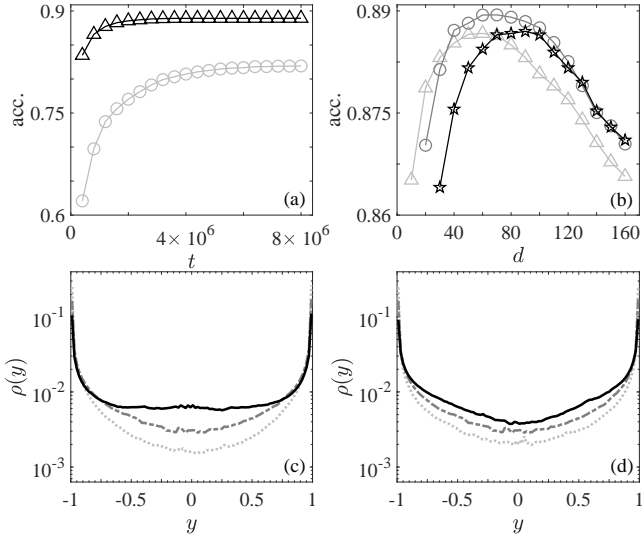}
  \caption{\label{fig:9}The mechanism of overtraining. (a) Test accuracy as a function of training time for the linear (circles) and nonlinear (triangles) neural networks. (b) Maximum test accuracy as a function of $d$ for $k=0.05$ (triangles), $0.15$ (circles) and $0.3$ (stars). (c) The distribution of outputs of hidden-layer neurons for $d=30$ (solid line), $70$ (dashed-dotted line) and $100$ (dotted line) with $k=0.15$. (d) The distribution of outputs of hidden-layer neurons for $k=0.05$ (solid line), $0.15$ (dashed-dotted line) and $0.3$ (dotted line) with $d=70$.}
 \end{figure}

 How to maximally extract the linearly separable and inseparable features is an essential problem for obtaining the optimal neural network. Figure 9(b) shows the test accuracy of a three-layer nonlinear network of $N_1=600$ for several values of $k$ as a function of $d$. Here, the accuracy at a given value of $d$ is obtained similarly to those shown in Fig.9(a), that is, the neural networks are trained to reach the maximum training accuracy by increasing the training time. We see that, for a given $k$, increasing $d$ increases the accuracy initially till it reaches the maximum value, and further increasing $d$ actually decreases the accuracy, which is a sign of overtraining. Similarly, at a fixed $d$, increasing $k$ increases the accuracy initially, but further increasing causes overtraining and the accuracy to decrease. It seems that the understanding and controlling of overtraining is the key to finding the optimal solution.

 Studying the distribution of outputs from hidden-layer neurons, we find that the overtraining induced by excessive $d$ or $k$ is all due to the presence of excessive number of neurons working in the extremely nonlinear regions with $y=f(\cdot) \approx \pm 1$. In Fig. 9(c) and 9(d) we show the distribution of outputs of hidden-layer neurons with $d = 30,70,100$ and $k = 0.15$, and with $k = 0.05,0.15,0.3$ and $d = 70$, respectively. In both plots, the distributions evolve from the state with less extremely outputs (with more outputs around $y = 0$) to the state with more extreme nonlinear outputs (with more neurons close to $y \approx \pm 1$ and characterised by the hight of the two peaks) as $d$ increases from $30$ to $100$ in Fig. 9(c) or as $k$ increases from $0.05$ to $0.3$ in Fig. 9(d). We see that the maximum accuracy appears with a proper distribution.

 Neurons working in the extreme nonlinear regions are necessary in the extracting of linearly inseparable features. Similar to the mechanism shown by Fig.~\ref{fig:5}(a) and 5(b), different classes of hidden-layer neurons working in the extremely nonlinear region are necessary for constructing the nonlinear learning mode. However, extremely nonlinear neurons do have their drawbacks; they decrease the robustness of the neural network since they behave like the step neurons, $f(\cdot) = {\rm sign}(\cdot)$, and too many such neurons would lead to the off-balance between the extraction of linearly separable and inseparable features. In more detail, for a given training set, the proportion of linearly separable and inseparable features of samples is fixed. With a fixed number of hidden-layer neurons, there are two possible scenarios that may lead to a less effective neural network. One is that the total number of neurons is insufficient for extracting the complete information. Another is that the ratio of neurons with linear and nonlinear learning modes does not match that of the linearly separable and inseparable features.

 With this understanding we can explain the results of Fig.9(b). Because there is a plenty of linearly separable features, the neural network tends to invoke the linear learning mode to extract linearly separable features. At smaller  $k$ and $d$, the training goal is achieved by the linearly separable features alone, the test accuracy would be low because of losing the information contained in linearly inseparable features. With the increase of $k$ and $d$, nonlinear neurons with extreme nonlinearity increase, leading to more nonlinear learning mode triggered and linearly inseparable features extracted, and thus the accuracy is improved. However, excessive number of extremely nonlinear neurons would lead to the decrease of neurons with linear learning modes, and decrease the extraction of linearly separable features, and leads to overtraining.

 \begin{figure}[htp]
  \centering
  \includegraphics[width=1\linewidth]{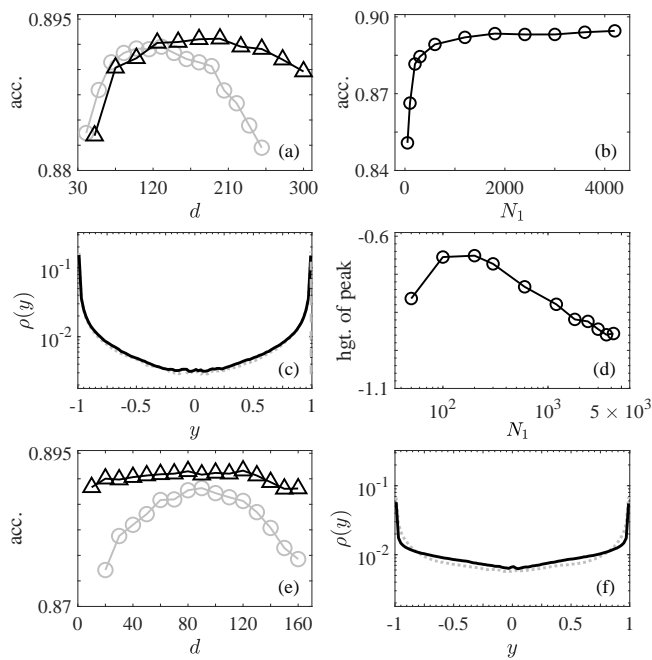}
  \caption{\label{fig:10} The role of width and depth on network performances, (a) to (d) are for three-layer networks, and (e) and (f) are for the four-layer and five-layer neural networks with the width fixed at $N_1=600$. In more detail, (a) shows the maximum test accuracy as a function of $d$ for $N_1=1200$ (gray circle) and $N_1=1800$ (black triangle), (b) shows the maximum test accuracy as a function of $N_1$, (c) shows the distribution of outputs of hidden-layer neurons for $N_1=1200$ (gray-dashed curve) and $N_1=1800$ (black-solid curve) at the maximum test accuracies, (d) shows the height of the picks of the distribution as a function of $N_1$ with semilogarithmic scale, (e) sows the maximum test accuracy as a function of $d$ for the four-layer (gray circle) and five-layer (black triangle) neural networks, and (f) shows the distribution of outputs of hidden-layer neurons for the four-layer (gray-dashed curve) and five-layer (black-solid curve) neural networks.}
 \end{figure}

\subsubsection{Increasing the width or depth of neural networks can maximally extract the linearly separable and inseparable features}\label{}

Increasing the width and the depth of neural networks provides two effective ways to maximally extract both the linearly separable and inseparable features. To show the width effect, we study a 784--${\rm N_1}$--10 neural network trained by $600$ MNIST samples with $k = 0.15$, which is the optimal value for $N_1 = 600$ as shown in the last subsection; it is checked that this value remains optimal for $N_1$s used in this subsection. In Fig.~\ref{fig:10}(a), the circles and triangles show the accuracy as a function of $d$ with $N_1=1200$ and $N_1=1800$, where each data point is the highest accuracy achieved at each value of $d$. Fig. ~\ref{fig:10}(b) shows the optimal accuracy as a function of the width where the optimal $d$ is searched and used in its calculation. In Fig.~\ref{fig:10}(c) we show the distribution of the outputs of hidden-layer neurons for $N_1=1200$ and $1800$. In Fig. ~\ref{fig:10}(d) we show the average height of the two peaks of the distribution as a function of the width $N_1$.

We see from Fig.~\ref{fig:10}(a) and ~\ref{fig:10}(b) that the accuracy increases with the increase of width and gradually tends to saturation around $N_1=1500$, implies that with a sufficient width, the neural network has extracted both linearly separable and inseparable features. Fig.~\ref{fig:10}(c) and ~\ref{fig:10}(d) indicate that the proportion of neurons with extremely nonlinearity decreases while linear neurons with outputs around $y =0$ increases with the increase of width for large enough $N_1$. In the region of small $N_1$, Fig.~\ref{fig:10}(d) indicates that the neurons with extremely nonlinearity increases initially while turns to decrease after a critical $N_1$, and is scaled as $\sim 1/N_1$ when $N_1$ is large enough. These two plots reveal how the width play the role. For a relatively small-sized neural network, since the linearly separable features are dominant in the MNIST samples, the network is inclined to use more neurons to extract this kind of information, leading to neurons with extremely nonlinearity has a relatively small ratio. In such a case, the neural network has a low accuracy due to lock of neurons to extract completely both linear and nonlinear features. With the increase of the width, there are more neurons available and thus some of them can be spared to extract the linearly inseparable features, and thus the ratio of extremely nonlinear neurons can increase to a more optimal value. However, if $N_1$ is large enough, with the optimal parameters the total number of the nonlinear neurons applied for extracting the linearly inseparable feature may become saturated. This statement can be inferred from the fact that the height of peaks decreases as  $1/N_1$. Note that the heights close to $y = \pm 1$ represent the proportion of neurons with extremely nonlinearity. Only when the number of these neurons keeps as a constant, its proportion may scale as $cost./N_1$. The reason that the extremely nonlinear neurons are maintained without further increasing should be that neurons with extremely nonlinearity are harmful to the network performance due to having large input-output sensitivity, and the formation of the nonlinear learning mode needs a combination process among different neurons, as has been pointed out early.

To show the depth effect, we study a four-layer 784--600--600--10 neural network and a five-layer 784--600--600--600--10 neural network trained by the first 600 MNIST samples. In the $8$th and $9$th rows of Fig.~\ref{fig:7}, we show the radiographs of the ten output neurons of these two neural networks, which indicate that the ``holographic'' structure of the network is remained. Figure~\ref{fig:10}(e) shows the accuracy of these two neural networks as a function of $d$. Together with the result of the three-layer neural networks shown in Fig.~\ref{fig:9}(b), we see that the test accuracy increases with the depth.

We find an important superiority of a deeper-layer neural network over a wider-layer one for practical applications. From Fig. ~\ref{fig:9}(b) and Fig.~\ref{fig:10}(a) we see that the optimal value of $d$ (with maximum test accuracy) increases with the increasing width quite significantly. To obtain the optimal test accuracy of a much wider neural network, one would have to search in a wide range for the optimal $d$ (For neural networks using the softmax cost function, this means one requires a long enough training time). In contrast, as shown in Fig. ~\ref{fig:9}(b) and Fig.~\ref{fig:10}(e), the optimal values of $d$ are approximately identical for three-layer to five-layer neural networks. More importantly, there is a tendency that the the test accuracy turns to $d$-insensitive with the increase of the depth. Particularly, in the case of the five-layer neural network, a wide range of $d$ that would give approximately the optimal test accuracy. It implies that once the goal of training with a small $d $ is achieved,  the neural network has already gain the optimal test accuracy approximately. As a result, it would be much easier to obtain the optimal solution for a deep-layer neural network.

Figure ~\ref{fig:10}(f) shows the output distribution of neurons in the fist hidden layer. It indicates that the number of extremely nonlinear neurons decreases with the increase of depth, implying that the mechanism is similar to that with the increasing of the width. From the point of view of weight pathways, a deeper neural network has much more weight pathways comparing to a shallower one even with the same number of hidden-layer neurons. For example, a 784--1200--10 network has $784 \times 1200 \times 10 \sim 9 \times 10^6$ weight pathways, while a 784--600--600--10 network has $784 \times 600 \times 600 \times 10 \sim 3 \times 10^9$ weight pathways. Therefore, deep-layer networks have much more freedom to establish subnetworks, and should provide more freedom to construct learning modes. This should be the reason that a deep-layer neural network can have the nonlinear learning mode with a small $d$ to extract the linearly inseparable features. The details of this mechanism are still to be investigated in the future.

\section{SUMMARY AND DISCUSSIONS}\label{sec:5}

 Weight pathways connect inputs to outputs of a neural network through subnetworks that are characterized by penetration coefficients; any internal change of the network can be qualitatively displayed by the changes of such penetration coefficients. Therefore, the WPA approach provides effective means to detect the internal structure of the network. In more details, with hidden-layer neurons as nodes, the network can be decomposed into a series of subnetworks, each of which is characterized by a characteristic map with a set of penetration coefficients. The training process can be interpreted as the organization of weight pathways to produce the minimum cost function, creating coherence structures for the penetration coefficients that represent the enhancement or suppression on the corresponding component of the input vector. With the visualization of the penetration coefficients, we gain a penetrating view of the inside of the ``black box'', i.e., the neural network, making its learning and recognition mechanisms interpretable. One of the important findings through the visualizations of the penetration coefficients is that each subnetwork uses a ``holographic'' structure to encode all the training samples instead of just one class of the samples. The ``holographic'' structure reveals how recognition is performed, that is, when an input vector is presented, every subnetwork reacts to the whole information in it with its ``holographic'' structure and decides collectively what the input vector is or is not. The ``holographic'' structure is also the basis of the generalization capability of the neural network. Our findings support the latest neurobiological understanding of the biological neural networks in the brain~\cite{barrett2020seven,doi:10.1126/science.aax1512}, and reveals that information is stored in the networks of weight pathways.

 The WPA approach reveals the self-organizing differentiation of the hidden-layer neurons. After the self-organization, hidden-layer neurons can be generally divided into dominant neurons and auxiliary neurons. The former realizes the goal of classification, while the latter helps to approach the goal of training. The analysis of the differentiation of the hidden-layer neurons leads to the finding of the linear and the nonlinear learning modes. The former extracts linearly separable features and can be performed by single class neurons independently. The latter can extract linearly inseparable feature and requires the cooperation of multiple classes of neurons and the help of the nonlinearity of the neuron transfer function.

 The WPA approach not only reveals what to learn and how to learn, but also reveals how to learn better. What to learn and how to learn --- extract the linearly separable features with linear learning mode, and extract linearly inseparable features with nonlinear learning mode. How to learn better --- maximize the extraction of both linearly separable and linearly inseparable features. Our study shows that a neural network tends to first extract linearly separable features with the linear learning mode. If each sample contain sufficient linearly separable features, the network may reach the training goal relying only on the linear learning mode, thus wasting linearly inseparable feature.

 In order to extract the linearly inseparable features, the network needs use a cost function that is difficult to reach the true minimum with only the linearly separable features, and then the network will invoke the nonlinear learning modes. To maximally utilize the information of the samples, the neural network must extract both features completely and with the right balance. We demonstrate that increasing the width and depth of the network are effective strategies for this purpose. When the width of a network is large enough, there are enough neurons to support both the linear and nonlinear learning modes, and thus one has a good chance to extract both features completely. Increasing the depth of the network can also increase the probability of nonlinear learning mode, and improve the network performance. More importantly, we find that increasing the depth of a network can get the optimum performance relatively easier than increasing the width. In more details, the optimum control parameters stay approximately constant for different depths, hence the careful search for the control parameters is not necessary. This property is very beneficial for applications. We have not studied in detail neural networks with deeper layers, other types of neuron transfer function, other updating algorithms, or other types of networks. We hope such studies can be shown in future.

\section*{Acknowledgment}

We acknowledge support by the NSFC (Grants No. 11975189, No. 11975190).

%


\bibliography{mec}



\end{document}